# A New Spatiotemporal Correlation Anomaly Detection Method that Integrates Contrastive Learning and Few-Shot Learning in Wireless Sensor Networks

Miao Ye[a,c,f]*, Suxiao Wang[b,d], Jiaguang Han[b,d], Yong Wang[a,e,f], Xiaoli Wang[a,c], Jingxuan Wei[a,c], Peng Wen[a], Jing Cui[e,f]

[a] Ministry of Education Key Lab of Cognitive Radio and Information Processing, Guilin University of Electronic Technology, Guilin, China,

[b] College of Precision Instrument and Optoelectronic Engineering, Tianjin University, Tianjin, China,

[c] School of Computer Science and Technology, Xidian University, Xi'an, Shaanxi, China

[d] Guangxi Key Laboratory of Optoelectronic Information Processing, School of Optoelectronic Engineering, Guilin University of Electronic Technology, Guilin, China

[e] School of Computer Science and Information Security, Guilin University of Electronic Technology, Guilin, China

[f] Guangxi Engineering Technology Research Center of Cloud Security and Cloud Service, Guilin University of Electronic Technology, Guilin, China

ARTICLE INFO

ABSTRACT

Detecting anomalies in the data collected by Wireless Sensor Networks (WSNs) can provide crucial evidence for assessing the reliability and stability of WSNs. Existing methods for WSN anomaly detection often face challenges such as the limited extraction of spatiotemporal correlation features, the absence of sample labels, few anomaly samples, and an imbalanced sample distribution. To address these issues, a spatiotemporal correlation detection model (MTAD-RD) considering both model architecture and a two-stage training strategy perspective is proposed. In terms of model structure design, the proposed MTAD-RD backbone network includes a retentive network (RetNet) enhanced by a cross-retention (CR) module, a multigranular feature fusion module, and a graph attention network module to extract internode correlation information. This proposed model can integrate the intermodal correlation features and spatial features of WSN neighbor nodes while extracting global information from time series data. Moreover, its serialized inference characteristic can remarkably reduce inference overhead. For model training, a two-stage training approach was designed. First, a contrastive learning proxy task was designed for time series data with graph structure information in WSNs, enabling the backbone network to learn transferable features from unlabeled data using unsupervised contrastive learning methods, thereby addressing the issue of missing sample labels in the dataset. Then, a caching-based sample sampler was designed to divide samples into few-shot and contrastive learning data. A specific joint loss function was developed to jointly train the dual-graph discriminator network to address the problem of sample imbalance effectively. In experiments carried out on real public datasets, the designed MTAD-RD anomaly detection method achieved an F1 score of 90.97%, outperforming existing supervised WSN anomaly detection methods while also achieving shorter inference times.

## 1. Introduction

Wireless sensor networks (WSNs) are network systems composed of many distributed sensor nodes. These nodes collect, process, and transmit data via wireless communication (Jamshed, Ali, Abbasi, Imran, & Ur-Rehman, 2022). Each network node has certain sensing, computing, and communication capabilities, enabling it to monitor and measure diverse environmental parameters and generate time series data. Owing to their numerous nodes and wide coverage areas, WSNs offer high flexibility and scalability, rendering them an optimal solution for real-time monitoring and data collection. They have been widely applied in various fields, such as environmental monitoring (Giri, Ng, & Phillips, 2019), industrial inspection (Lee, & Ke, 2018), and intelligent transportation (Abbas, & Yu, 2018).

In general, the data collected by WSNs are considered multidimensional time series data. Owing to various interference factors, the data collected by WSNs may display anomalous deviations. These anomalies can result not only from security threats but also from sensor node failures or anomalous events occurring within the deployment area. WSN data anomalies are typically defined as data distributions that deviate significantly from the normal observed data distribution. When the distribution characteristics of WSN time series data solely along the temporal dimension are considered, WSN data anomalies are categorized into point anomalies, collective anomalies, and contextual anomalies (OReilly, Gluhak, Imran, & Rajasegarar, 2014; Chandola, Banerjee, & Kumar, 2009). Specifically, a point anomaly refers to a sharp deviation in the data at a particular time point, where this data point varies greatly from the surrounding data at adjacent time points. A collective anomaly refers to a set of data points collected over a time interval that deviate from the normal data pattern. A contextual anomaly occurs when data at a specific time or time interval conform to the normal data pattern but do not align with the data patterns observed in the preceding or succeeding time intervals. When considering the spatial and temporal correlation distribution characteristics of data collected by multiple WSN nodes, the data anomalies of WSN nodes also include the anomaly that the correlation distribution of WSN time series data deviates from the normal correlation distribution (Ding, Yu, & Wang 2020; Li, & Jung, 2023). For example, temperature time series data and humidity time series data from the same sensor node will have a stable negative correlation with a long-term trend. When the correlation between the two physical quantities suddenly becomes irrelevant or positive, the correlation between the temperature time series data and humidity time series data is anomalous. In addition to the existence of correlation anomalies within time series data collected by the same sensor node, it is also possible for such correlation anomalies may also arise between time series data collected by different sensor nodes. For example, when a fire event happens, the temperature data collected by different sensor nodes around the fire source show the same trend. This indicates that the different temperature time series data among these different nodes are positively correlated. When the temperature time series data of one node violates the change rule of this positive correlation, we believe that the spatial correlation of the time series data of this node appears to be anomalous. In real-world scenarios, such data anomalies, which reflect abnormal system states, can impair the overall reliability and stability of WSNs. Therefore, anomaly detection in wireless sensor network nodes is of great practical significance.

Traditional anomaly detection methods, such as autoregressive (AR) models (Jung, Berges & Poczos, 2015) and autoregressive integrated moving average (ARIMA) models (Yu, Jibin & Jiang, 2016), rely primarily on statistical features for feature extraction and assume linear relationships in the data. However, these methods exhibit poor performance when applied to high-dimensional, nonlinear data from WSNs, and they are highly susceptible to noise. As a result, machine learning techniques have gradually been introduced into this field. Methods such as support vector machines (SVMs) (Hosseinzadeh, Rahmani, Bidaki, Masdari & Zangakani, 2021) and clustering (Liu, Gao, Wen & Luo, 2023) have demonstrated the ability to capture potential anomalies within more complex data patterns, significantly improving detection accuracy. However, machine learning approaches still face challenges such as the "curse of dimensionality," difficulties in feature selection, and insufficient modeling of long-term dependencies. With the advancement of deep learning technologies, anomaly detection methods based on convolutional neural networks (CNNs) (Sarangi, Mahapatro & Tripathy, 2021), recurrent neural networks (RNNs) (Lazar, Buzura, Iancu & Dadarlat, 2021), and long short-term memory (LSTM) networks (Matar, Xia, Huguenard, Huston & Wshah, 2023) have been proposed. These methods are proficient in capturing nonlinear relationships in WSN data and learning more complex feature representations, thereby greatly enhancing the ability to handle high-dimensional data. The emergence of transformer models has further addressed the issues of gradient explosion and gradient vanishing when handling long-term dependencies in WSN anomaly detection tasks (Tuli, Casale & Jennings, 2022; Wu, Xu, Wang, & Long, 2021). In retentive networks (RetNets), the attention mechanism in transformers has been replaced with a retention mechanism, enabling RetNet-based models to perform inference serially on the basis of data at each time step, ensuring that the model's inference cost does not increase as the sample sequence length increases (Sun, Dong, Huang, Ma, Xia, Xue, Wang & Wei, 2023; Long, Zhu & Xiao, 2024). To address the challenge of extracting spatial correlation features in WSN anomaly detection, graph neural networks (GNNs) have been applied to the detection of anomalous nodes in WSNs (Wang & Liu, 2024; Yan, Luo & Shao, 2023). GNN-based methods not only capture spatial information between WSN nodes but also treat different modalities of data as nodes in a complete graph, thus extracting correlation information between these modalities. This explicit handling of correlations among multidimensional time series data in WSNs enhances the interpretability of the models. However, anomaly detection methods based on transformers and GNNs still have the following shortcomings in WSN applications:

First, the existing WSN anomaly node detection methods based on deep learning only consider different timing modes in the same WSN node (Zeng, Chen, Qian, Wang, Zhou & Tang, 2023) or the correlation between the same timing modes between different nodes (Zeng, Chen, Qian, Wang, Zhou & Tang, 2023; Dohare & Tulika, 2021) when processing the correlations between WSN high-dimensional time series data. The spatiotemporal correlation between multiple temporal modes of different WSN nodes is rarely considered, which limits the performance of WSN anomaly detection.

Second, owing to the complex spatiotemporal correlations and inherent long-term dependence mentioned above between WSN time series data, it is difficult to label WSN data samples, which unavoidably gives rise to issues of missing labels for anomalous data. Supervised learning-based WSN anomaly detection methods are unable to utilize these unlabeled data effectively, limiting their ability to capture the true data distribution and consequently impacting model generalization performance (Ahasan, Haque & Alam, 2022). Several studies have employed reconstruction-based approaches (Luo, & Nagarajan, 2018; , Xu, Wang, Jian, Liao, Wang & Pang, 2022) to address the issue of missing labels in WSN data. However, reconstruction-based methods compress and reconstruct WSN time series data at the structural level, emphasizing the global pattern features of the samples. This makes them less effective in identifying local single-point anomalies or short-term fluctuations in data that otherwise conform to normal overall trends, as the reconstruction error in such cases is often negligible, leading to potentially missed detections. Furthermore, the effectiveness of reconstruction-based anomaly detection methods is highly dependent on the quality of the WSN data. When the training set contains anomalous data, the model's ability to extract the true data distribution is constrained.

Finally, the performance of the WSN anomaly node detection method is closely related to the number of samples and the sample distribution, but WSN datasets generally face the problem of sample imbalance due to the limited number of anomalous samples. Several studies have addressed these issues by using resampling techniques to mitigate the imbalance between normal and anomalous samples. However, oversampling the minority class of anomalous samples might lead to severe model overfitting, whereas undersampling the majority class of normal samples may result in the loss of critical information (Huan, Lin, Li, Zhou & Wang, 2020; Sun, Wang, Tang & Bi, 2023). Other studies have used generative adversarial networks (GANs) to generate additional samples for underrepresented classes (Di Mattia, Galeone, De Simoni & Ghelfi, 2019; Miao, Jin, Yuan, Qiuxiang & Yong, 2024). However, GANs rely on learning the overall distribution of the original data to generate realistic samples. When the number of anomalous samples in the original dataset is inadequate, GANs may struggle to capture the characteristics of anomalous data accurately, leading to generated samples that deviate from the true distribution of anomalies. This inconsistency can compromise the detection accuracy of subsequent discriminative networks. Therefore, mitigating the impact of insufficient anomalous samples on model training remains a vital consideration when WSN anomaly node detection methods are designed.

To address the challenges in anomaly detection for WSN time series data, including the difficulty in effectively detecting correlation anomalies among multiple nodes and modalities, the issue of missing sample labels due to complex spatiotemporal correlations, the scarcity of anomalous samples, and class imbalance, this paper proposes a novel anomaly node detection network. The proposed network, named multitemporal anomaly detection based on RetNet and dual graph networks (MTAD-RD), is designed from two perspectives—the model architecture and training strategy—facilitating the integration of multinode and multimodal features for enhanced detection performance.

First, to efficiently capture the latent internode and intermodal correlations in WSN temporal data, a cross-retention (CR) block is designed on the basis of RetNet (Sun, Dong, Huang, Ma, Xia, Xue, Wang & Wei, 2023) to extract the intermodal correlations. This CR block is integrated with a graph attention network (GAT), which is utilized to extract internode correlation information, forming the backbone network of the proposed MTAD-RD. The designed backbone network effectively exploits spatiotemporal correlations, supports parallel training and sequential inference, and reduces inference-phase computational overhead. Additionally, a multigranular feature fusion module is incorporated into the backbone network, augmenting the representation of WSN node features. Second, to make use of the unlabeled data in the dataset for model training, a contrastive learning proxy task customized to the characteristics of WSN temporal data is designed. This task is based on node-to-subgraph contrast and accomplishes two objectives: it incorporates first-order subgraphs into subgraphs constructed using a random walk method, thereby preserving spatial correlations, and it defines positive and negative sample pairs on the basis of the similarity between anchor nodes and subgraphs. As a result, the pretraining process of the MTAD-RD backbone network does not rely on manually annotated information. Finally, to address the issue of the limited number of anomalous samples and the resulting class imbalance, this paper integrates few-shot learning with contrastive learning. A cache-based sampler is designed to partition the output features of the backbone network into a support set and query set for few-shot learning, as well as into anchor samples and positive–negative sample pairs for contrastive learning. By designing a correlation joint loss function, the multistrategy joint optimization of the anomaly node discriminant network in a weakly supervised way based on a dual graph neural network is proposed.

In summary, this paper proposes an anomaly detection model for WSN multinode and multimodal temporal data and develops a two-stage training method to address various challenges effectively, such as the insufficiency of labeled training data, the scarcity of anomalous samples, and class imbalance. The main contributions of this work are as follows:

1) Compared with existing WSN anomaly detection methods, which are only applicable to scenarios with the different timing modes in the a same WSN node or the same timing modes between different nodes, this paper designs a backbone network that integrates spatiotemporal correlation information both between WSN nodes and across modalities, building on the improved RetNet architecture. Furthermore, the backbone network of the proposed MTAD-RD model supports sequential inference, significantly reducing inference overhead.

2) To address the issue of missing label information in WSN datasets, we design an unsupervised contrastive learning proxy task, leveraging the graph-structured characteristics of WSN temporal data. The MTAD-RD model's backbone network is pretrained with a vast amount of unlabeled data to gain transferable knowledge, resolving the problem of high labeling costs for WSN samples.

3) To address the issue of the limited number of anomalous samples and the consequent class imbalance in WSN datasets, this paper integrates few-shot learning with contrastive learning, and a cache-based sample sampler and a joint loss function are proposed to perform weakly supervised joint optimization of the anomaly node discriminant network in the proposed MTAD-RD model. This approach effectively alleviates the negative effects of insufficient anomalous samples and sample imbalance on model training.

The remainder of this paper is organized as follows: Section 2 introduces existing work and research achievements in the field of WSN anomaly node detection; Section 4 provides the mathematical definitions relevant to the research in this paper; Section 4 presents a detailed introduction to the various components and optimization strategies of the MTAD-RD model; Section 5 presents the results obtained via the proposed method on real data and compares them with the results obtained by existing methods; and Section 6 summarizes the content of this paper and discusses future research directions.

## 2. Related work

In recent years, deep neural networks have been applied extensively in the field of anomaly detection because of their powerful ability to model temporal dependencies. Since the two fundamental components of deep learning methods—model structure and data—are essential for deep neural networks, the anomaly detection work presented in this paper concentrates on both aspects. Therefore, this section introduces relevant work on model design and training data for anomaly detection, including how existing approaches design WSN anomaly detection models on the basis of spatiotemporal correlations in high-dimensional temporal data, as well as how anomaly detection studies have addressed issues such as the lack of sample labels, the scarcity of anomalous samples, and class imbalance, by proposing corresponding training strategies.

Anomaly detection is a classic task in machine learning, with feature extraction serving as a critical core step. Effective feature extraction methods can extract essential characteristics from vast amounts of raw data, enabling the classification of anomaly types. Traditional anomaly detection methods rely primarily on manually designed features and rules for feature extraction. For example, distance-based anomaly detection methods evaluate anomalies by measuring the similarity between samples using metrics such as the Euclidean distance, Manhattan distance, and cosine similarity (Du, Li, Zhao, Jiang, Shi, Jin & Jin, 2024; Pang, Cao, Chen & Liu, 2018). Deep learning methods based on deep neural networks, which can automatically and comprehensively learn features, have emerged as research hotspots because of their ability to address complex anomaly detection tasks effectively. Since WSN data are inherently time series data, many studies have investigated the application of deep learning techniques to extract temporal features from WSN time series data. For example, (Ullah & Mahmoud, 2022) incorporated recurrent connections to capture the temporal dependencies within WSN time series data. However, as the number of time steps increases, the gradients during backpropagation may diminish or explode, leading to vanishing or exploding gradient problems. This limitation hampers RNN-based feature extraction models from effectively learning long-term dependencies. To address these challenges, (Wei, Jang-Jaccard, Xu, Sabrina, Camtepe & Boulic, 2022; Tang, Xu, Yang, Tang & Zhao, 2023) proposed the use of LSTM and GRU networks as alternatives to RNNs for temporal feature extraction. These architectures incorporate gating mechanisms to mitigate the gradient vanishing and explosion issues that are encountered when capturing long-range dependencies. Furthermore, the success of transformer models in the field of natural language processing (NLP) has inspired new approaches for temporal feature extraction. For example, (Chen, Chen, Zhang, Yuan & Cheng, 2021) employed the transformer as a temporal extraction network, entirely discarding the recurrent structure of RNNs. By capitalizing on the self-attention mechanism, the designed model can capture temporal information between any two time points in a single computation, significantly improving training efficiency, parallelization, and the modeling of long-term dependencies. Inspired by the sparse self-attention mechanism, (Ding, Zhao & Sun, 2023) proposed the square-root sparse self-attention mechanism to address the computational overhead tied to transformers when processing long time series data. This method, combined with local self-attention, enables the model to leverage more proximate data at each time step for prediction, thereby avoiding the loss of local information. However, WSN anomaly detection methods based on sparse self-attention mechanisms often introduce additional hyperparameters, and an inappropriate choice of sparsification strategy may result in the loss of critical temporal information, adversely affecting the model's detection accuracy. With the progression of large language models (LLMs), new models for extracting long-sequence features have been proposed. For example, in RetNet, the attention mechanism in transformers is replaced with a retention mechanism to reduce inference overhead (Sun, Dong, Huang, Ma, Xia, Xue, Wang & Wei, 2023; Long, Zhu & Xiao, 2024). During the training phase, RetNet-based models, similar to transformer-based models, can process data from all time steps in parallel. However, in the inference phase, RetNet-based models perform serial inference on data from each time step, ensuring that inference costs remain invariant with respect to the sequence length. Despite these advantages, these methods are limited to extracting temporal correlation features from WSN time series data without capturing the spatial correlations inherent to WSN data with graph structures. Consequently, they fail to perform spatial correlation anomaly detection on WSN time series data.

To comprehensively extract spatiotemporal correlation features from multidimensional time series data with graph-structured characteristics in WSNs, recent studies have explored methods based on GNNs and attention mechanisms. GNNs are well suited for representing graph-structured data and can capture both local and global dependency relationships between nodes through edge information. For example, (Ding, Sun & Zhao, 2023) proposed MST-GAT, which employs a GAT to model the correlations between different sensor nodes. By incorporating a graph attention module, the model captures the spatial correlations among nodes, enabling it to represent internode relationships effectively. In addition, attention mechanisms integrate temporal features by calculating the similarity between different feature sequences. (Li, Yang, Wan & Li, 2022) designed two transformer-GRU models based on attention mechanisms to capture the long-term dependencies in time series data and the correlations among different temporal features. Specifically, the F-transformer-GRU treats features from different time sequences as word vectors and calculates self-attention coefficients to infer relationships between sequences. (Bai, Wang, Zhang, Miao & Lin, 2023) proposed CrossFuN, which is a novel anomaly detection architecture designed under the assumptions of temporal-frequency heterogeneity and temporal-frequency coordination. CrossFuN employs a temporal-frequency joint cross-fusion block to capture interrelationships among time series from the perspective of both the temporal and frequency domains. However, these approaches focus primarily on extracting spatiotemporal features from a single modality of time series data across multiple WSN nodes. They do not address the extraction of relational features among multiple modalities of time series data across various WSN nodes.

The development of efficient deep learning-based WSN anomaly detection models also requires high-quality data. However, in practical WSN data collection scenarios, the data often exhibit characteristics such as high dimensionality, substantial redundancy, and high noise levels. Additionally, challenges such as missing data labels, a scarcity of anomalous data samples, and imbalanced class distributions further complicate the task. Many researchers have employed techniques such as reconstruction methods, contrastive learning, and few-shot learning to increase the performance of models. To address the issue of missing labels in WSN data, models are frequently designed on the basis of reconstruction methods, with the training objective set to minimizing the discrepancy between the model's input and output. A pioneering study on applying reconstruction methods to time series anomaly detection (Malhotra, Ramakrishnan, Anand, Agarwal & Shroff, 2016) was the first to employ LSTM networks within an encoder–decoder architecture. This approach trains the model exclusively on normal data, which ensures that normal patterns are reconstructed with high fidelity during inference, resulting in low reconstruction errors. In contrast, anomalous data, which deviate from normal patterns, yield significantly larger reconstruction errors. Samples with reconstruction errors exceeding a predefined threshold are identified as containing anomalies. Researchers have further introduced predictive branches into reconstruction-based models. For example, (Zhao, Wang, Duan, Huang, Cao, Tong, Xu, Bai, Tong & Zhang, 2020) designed two parallel GAT layers within a reconstruction model to dynamically learn the relationships among different time series and timestamps. The learned latent features were then used to predict the data for the next time step. Similarly, (Han & Woo, 2022) integrated a sparse autoencoder (SAE) with GNNs, incorporating sparsity constraints during the extraction of latent features from high-dimensional time series data. Additionally, a predictive branch was introduced, where latent features are input into a GNN to forecast the data for the subsequent time step. By incorporating predictive branches into reconstruction-based models, (Zhao, Wang, Duan, Huang, Cao, Tong, Xu, Bai, Tong & Zhang, 2020; Han & Woo, 2022) improved the robustness and adaptability of these models. However, models designed with reconstruction methods are highly dependent on the quality of the dataset. If the training set contains anomalous samples or is heavily contaminated with noise, such a model's ability to learn normal patterns may be compromised, thereby restricting the effectiveness of reconstruction-based models in detecting anomalous samples.

As a self-supervised learning method within the realm of unsupervised learning, contrastive learning offers an alternative approach for addressing the issue of missing labels. This method aims to bring similar samples closer together in the feature space while pushing dissimilar samples farther apart. By designing appropriate proxy tasks, contrastive learning methods can extract transferable knowledge from large amounts of unlabeled data, which can then be applied to downstream tasks. In the context of graph anomaly detection, the methods proposed in (Zheng, Li, Li & Gao, 2019; Liu, Pan, Wang, Xiong, Wang, Chen & Lee, 2021) operate under the assumption that there are no anomalous connections among edges within graph samples. Consequently, all graph samples are treated as positive pairs. To construct negative samples, original edges and nodes are sampled using Bernoulli distributions or random walks to generate subgraphs. However, these proxy tasks are tailored specifically for graph-structured anomaly detection and are unsuitable for detecting node-level temporal anomalies. (Yang, Zhang, Zhou, Wen & Sun, 2023) applied a purely positive contrastive learning approach to detect temporal anomalies, employing a dual-branch attention mechanism to obtain feature representations of samples from two distinct perspectives. The study posited that, under feature reordering conditions, the consistency of feature representations derived from different branches would be greater for normal samples than for anomalous nodes. Thus, the degree of feature consistency could be used as a criterion for anomaly detection. However, this purely positive contrastive learning approach lacks the constraints provided by negative samples, which can lead the model to favor outputs with identical embeddings, which is a phenomenon known as model collapse.

To address the difficulties of insufficient anomalous samples and the imbalance between normal and anomalous samples in WSN datasets, anomaly detection methods based on few-shot learning have been proposed. These methods aim to swiftly learn feature representations from a small number of anomalous samples. Finn et al. presented the model-agnostic meta-learning (MAML) algorithm (Finn, Abbeel & Levine, 2017), which attains rapid model convergence by fine-tuning task-sensitive parameters during the network training process. Through a few iterations, the algorithm promptly adapts to new tasks. Rusu et al. proposed the Latent Embedding Optimization (LEO) algorithm (Rusu, Rao, Sygnowski, Vinyals, Pascanu, Osindero & Hadsell, 2018), which involves the concept of a low-dimensional latent space. By updating network parameters within this latent space through inner-loop optimization, the method transfers prior knowledge and representations learned from large-scale datasets to target domains using few-shot learning, thereby enhancing model performance and generalization. Long et al. proposed a meta-learning graph attention network based on meta-learning, which trains a meta-learner on a dataset with labeled samples spanning diverse categories and generalizes it to fault diagnosis tasks using a small number of labeled samples (Long, Zhang, Yang, Huang, Liu & Li, 2022). Similarly, Wang et al. developed a few-shot learning method based on meta-learning that employs a dual graph convolutional network. This approach constructs instance graphs and instance distribution graphs to propagate label information, enabling effective fault classification (Wang, Tong, Wang, Xu & Song, 2023). These methods illustrate that accurate detection of anomalous samples is achievable even with a scarce number of such samples through few-shot learning. Given the advantages of few-shot learning, this study adopts a comparable approach, incorporating supplementary enhancements to further enhance model robustness.

Based on the analysis of the above-mentioned work, this study aims to enable model training on datasets with missing labels and a limited number of outlier samples, while fully exploiting the spatiotemporal information within the data. To achieve this, a RetNet model, improved through the design of Cross-Retention blocks, and a two-stage training framework is designed to optimize the MTAD-RD model through multi-strategy means. In the first stage of training, an unsupervised contrastive learning method is employed, utilizing node-to-subgraph contrast to pretrain the backbone network with large amounts of unlabeled data. In the second stage, multiple training strategies are integrated, with a major emphasis on few-shot learning to address the scarcity of anomalous data in WSN datasets. Additionally, contrastive loss is incorporated into the loss of dual-graph discriminative network. This incorporation further restricts the distribution of sample features in the feature space, thereby fortifying the connection between upstream and downstream tasks.

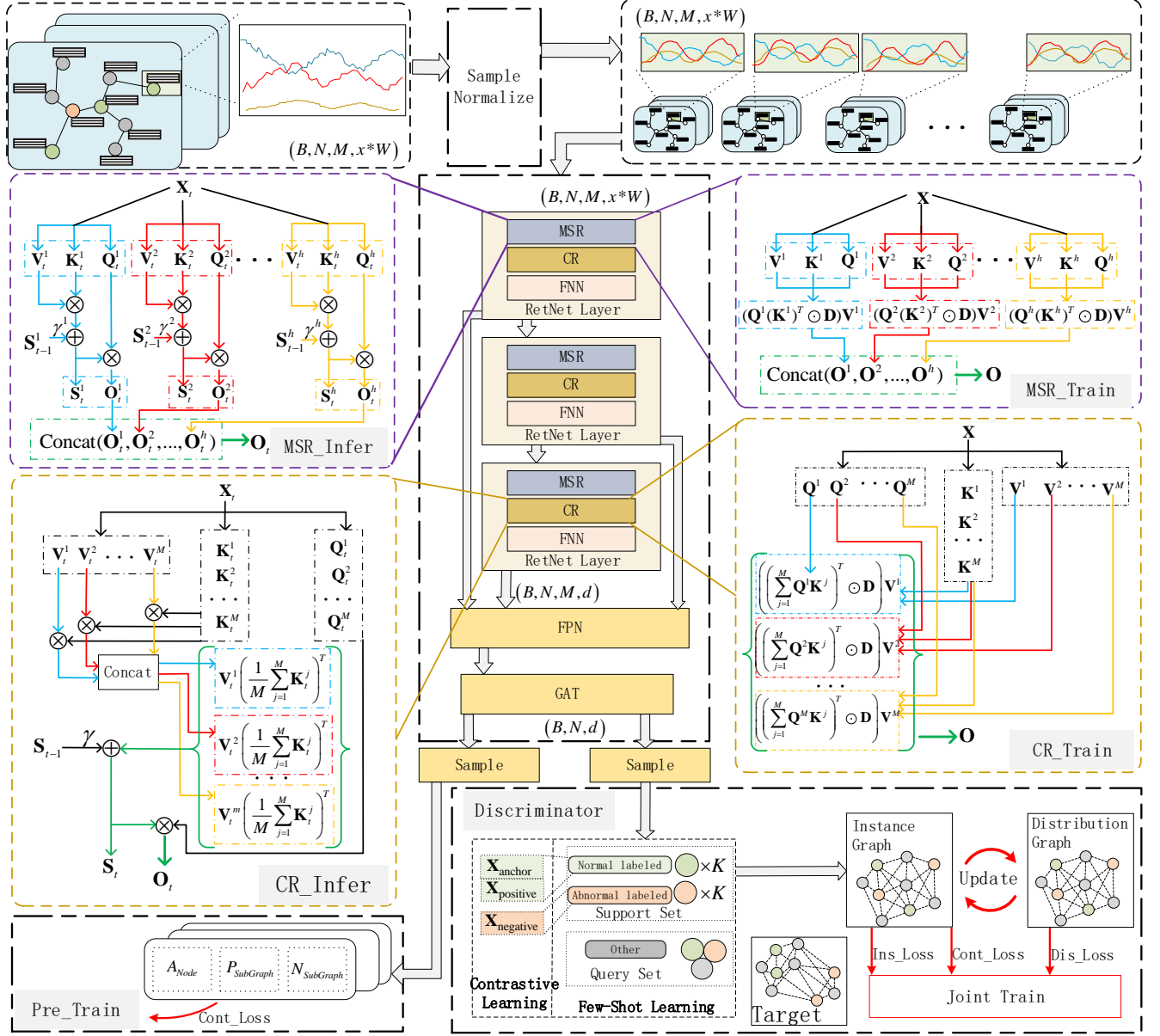

Fig. 1. Model Framework Diagram.

## 3. Problem definition

The data collected by a wireless sensor network at time $t$ can be represented as an attributed graph $G(t) = (A, X_t)$, where $A \in \mathbb{R}^{N \times N}$ represents the adjacency matrix of the attributed graph. $N$ is the number of sensor nodes, and $a_{i,j}$ is an element of the adjacency matrix $A$. There is a correlation between the data collected by the nodes of a wireless sensor network, which is determined by a combination of the distances between the nodes and the similarity of the deployment environment and is independent of the specific topology of the sensor network. The adjacency matrix A is used to indicate whether the correlation between sensor node $i$ and node $j$ needs to be considered. If there is a possible connection between node $i$ and node $j$, then $a_{i,j} = 1$; otherwise, $a_{i,j} = 0$. That is, only when $a_{i,j} = 1$ does the model consider the correlation between node $i$ and node $j$, and the specific coefficient of correlation is represented by the learnable parameters within the model. In general, after the deployment of the sensor network is complete, its spatial location does not change over time; thus, in this paper, the adjacency matrix A is considered not to change over time. $X_t \in \mathbb{R}^{N \times M}$ is the feature matrix, where $M$ is the number of physical quantities collected by a single sensor node (Jin, Koh, Wen, Zambon, Alippi, Webb, King & Pan, 2023). The data collected by the wireless sensor network from $t = 1$ to $t = T$ can be represented as $G[1:T] = (A, X) = \{G_1, G_2, ..., G_T\}$, where the feature matrix $X = \{X_1, X_2, ..., X_T\} \in \mathbb{R}^{N \times M \times T}$. Typically, the problem of anomaly node detection in wireless sensor networks can be formulated as determining a mapping function $F$.

$$y = F_\theta \left( G[t_1 : t_2] \right) \quad (1)$$

The term $G[t_1 : t_2]$ represents the attributed graph formed by the data collected by the WSNs from time $t_1$ to $t_2$. Typically, $F$ represents the mapping function corresponding to a neural network model (Zeng, Chen, Qian, Wang, Zhou & Tang, 2023; Ye, Zhang, Xue, Wang, Jiang & Qiu, 2024), and $\theta$ denotes the weight parameters of the neural network model associated with $F$ that need to be determined. $y \in \mathbb{R}^N$, where $y_i \in \{0,1\}$, is the output of the detection task. If $y_i = 1$, node $i$ is anomalous during the period from $t_1$ to $t_2$.

The symbols used in this paper and their meanings are shown in Table 1. The following sections provide a detailed introduction to each component of the design model.

Table 1

Symbols and definitions used in this paper

| Symbol | Definition |
|---|---|
| $B$ | The number of graph samples in a batch |
| $N$ | The number of nodes in the graph sample; the number of sensor nodes in WSNs |
| $M$ | The number of modes collected by the sensor node |
| $W$ | Sliding window size |
| $X$ | The attribute matrix corresponding to the graph sample |
| $X_t$ | The attribute matrix corresponding to time $t$ in $X$ |
| $W$ | Learnable parameters in the model |
| $Q$ | The query matrix obtained from the $X$ attribute matrix |
| $Q^i$ | The query matrix corresponding to the $i$-th mode in CR or the $i$-th header in MSR |
| $Q_t$ | The query matrix obtained from the data $X_t$ at time $t$ |
| $Q_t^i$ | The query matrix corresponding to the $i$-th mode or $i$-th header obtained from the data $X_t$ at time $t$ |
| $K$ | The key matrix obtained from the $X$ attribute matrix |
| $V$ | The value matrix obtained from the $X$ attribute matrix |
| $S_t$ | The state retention matrix obtained at the first time $t$ |
| $D$ | A lower triangular mask matrix representing the causal relationships in a sequence. |
| $\Theta$ | $\Theta$ represents the rotational encoding matrix implemented in RoPE (Su, Lu, Pan., Murtadha, Wen & Liu, 2021), and $\overline{\Theta}$ is its conjugate matrix. |
| $O$ | In each submodule, it represents the output matrix. |

Note: $K$, $V$, and $O$, like $Q$, use subscripts and superscripts to indicate different modalities or time steps. For example, $O_t^i$ represents the output derived from the attribute matrix $X_t$ at time step $t$ in the $i$-th head during MSR serialized inference. Learnable parameters in the model are often paired with subscripts and superscripts, such as $W_Q^i$, which denotes the parameter used to compute the query matrix in the $i$-th head.

## 4. Methodology

To leverage a lightweight model for capturing temporal information and potential spatiotemporal correlations in the WSN dataset, this paper proposes the MTAD-RD network model. The designed MTAD-RD model consists of three functional components: data preprocessing, a feature extraction network, and an anomaly detection network.

**Data Preprocessing:** This component is responsible for transforming the raw data collected from the WSNs into training samples for the model. The process includes downsampling, sample partitioning, and data normalization. When WSN data collection is excessively dense, downsampling significantly reduces the data volume and minimizes the impact of high-frequency noise. The sampling process also aligns the timestamps of different sensor nodes and repairs missing data.

**Feature Extraction Network:** This component is composed of an improved RetNet, a Graph Attention Network, and a feature fusion module. This paper presents the design of a CR block within the original RetNet framework. The CR enables RetNet to capture different inter-temporal relationships within a single sensor node when extracting inter-temporal dependencies. Furthermore, the feature fusion module with adaptive weights is employed to fuse sample features from different RetNet layers, allowing for the fusion of data with varying granularities. GAT adopts the graph attention mechanism, can effectively fuse the neighbouring node information of each node, and can avoid the over-smoothing problem that GCN may bring.

**Anomaly Detection Network:** The features and labels of each WSN node are organized into instance graphs and distribution graphs in the form of a complete graph. The node attributes and edge weights are updated in a coordinated manner through different multilayer perceptron (MLP) networks. In this study, we integrate few-shot learning with contrastive learning techniques to enhance the discriminative network through a joint loss function comprising instance graph classification loss, distribution graph classification loss, and contrast loss. Here, contrastive loss is used to constrain the feature distributions of normal and anomalous samples in both the instance and distribution graphs, whereas classification loss guides the final optimization direction of the detection network.

### 4.1. Data preprocessing

Sensor networks typically collect data that include noise. However, by appropriately increasing the time interval between data samples, the high-frequency fluctuations can be reduced, thereby mitigating the impact of noise. To minimize the impact of high-frequency noise and smooth the data, the original data are often downsampled. By increasing the sampling interval, each original sample can be divided into several new samples, allowing the periodic characteristics of the time series to be captured with a minimal temporal length. Therefore, this paper employs a sliding window approach with a step size of $k$ and a window size of $kW$ to segment the original data and perform downsampling to construct the samples. Assuming that the current time is $t$, the time series starting from $t$ with $kW$ data points is denoted as $X_{raw} = [x_t, x_{t+1}, ..., x_{t+kW-1}] \in \mathbb{R}^{M \times N \times kW}$. After downsampling, $X_{raw}$ is divided into $k$ samples $X_i \in \mathbb{R}^{M \times N \times W}$, where $i \in [1, k]$.

Since different modalities of data collected by WSNs have different dimensions; to eliminate the impact of dimensions and prevent model parameters from being dominated by data with larger dimensions, this paper standardizes the data of each modality via the Z score method. The standardization process is as follows:

$$X_{i,j}{}^* = \left(X_{i,j} - \mu(X_{i,j})\right) \big/ \sigma(X_{i,j}) \tag{2}$$

In formula (2), $X_{i,j}$ represents a time series of length $W$ corresponding to node $i$ and modality $j$ in sample $X$. $\mu(X_{i,j})$ denotes the mean of $X_{i,j}$, whereas $\sigma(X_{i,j})$ denotes the variance of $X_{i,j}$. $X_{i,j}^*$ is the normalized output sequence of $X_{i,j}$, where the data at each time step conform to a standard normal distribution.

*4.2. Temporal feature extraction module*

In the MTAD-RD model framework shown in Fig. 1, the MSR modules within each RetNet layer are designed to extract features over the temporal scale. Each MSR module consists of $h$ heads, which are multi-head retention blocks. These heads focus on different aspects, enhancing the model's feature representation. The MSR module has distinct forward computation processes during training and inference. During parallelized training, MSR operates similarly to multi-head attention in Transformer. It calculates the correlations between different time points via projection matrices $Q$ and $K$ and then integrates features across different time points on the basis of these correlations. In the serialized inference phase, features at time $t$ are derived from the retention coefficient matrix $S_t$ and the mapping matrix $Q_t$. The retention coefficient matrix $S_t$ at time $t$ is computed by adding the retention coefficient $K_t^T V_t$ at time $t$ to the historical retention coefficient matrix $S_{t-1}$. In light of the work presented in (Sun, Dong, Huang, Ma, Xia, Xue, Wang & Wei, 2023), the designed temporal feature extraction module can be represented as follows:

$$\begin{aligned}
&\boldsymbol{Q}^i = \boldsymbol{X}\boldsymbol{W}_Q^i \odot \boldsymbol{\Theta}, \boldsymbol{K}^i = \boldsymbol{X}\boldsymbol{W}_K^i \odot \bar{\boldsymbol{\Theta}}, \boldsymbol{V}^i = \boldsymbol{X}\boldsymbol{W}_V^i \\
&\boldsymbol{\Theta} = e^{it\theta}, \boldsymbol{D}_{t_1 t_2} = \begin{cases} \left(\gamma^i\right)^{t_1 - t_2}, & t_1 \geq t_2 \\ 0, & t_1 < t_2 \end{cases} \\
&\boldsymbol{O}^i = (\boldsymbol{Q}^i \boldsymbol{K}^{iT} \odot \boldsymbol{D})\boldsymbol{V}^i, i \in [1, h] \\
&\boldsymbol{O} = GN_h\left(\text{concat}\left(\boldsymbol{O}^1, \boldsymbol{O}^2, ..., \boldsymbol{O}^h\right)\right) \\
&\boldsymbol{Q}_t^i = \boldsymbol{X}_t \boldsymbol{W}_Q^i \odot \boldsymbol{\Theta}, \boldsymbol{K}_t^i = \boldsymbol{X}_t \boldsymbol{W}_K^i \odot \bar{\boldsymbol{\Theta}}, \boldsymbol{V}_t^i = \boldsymbol{X}_t \boldsymbol{W}_V^i \\
&\boldsymbol{S}_t^i = \gamma^i \boldsymbol{S}_{t-1}^i + \boldsymbol{K}_t^{iT} \boldsymbol{V}_t^i, \quad \boldsymbol{S}_0^i = \boldsymbol{0} \\
&\boldsymbol{O}_t^i = \boldsymbol{Q}_t^i \boldsymbol{S}_t^i, \quad i \in [1, h], t \in [1, W] \\
&\boldsymbol{O}_t = \text{concat}\left(\boldsymbol{O}_t^1, \boldsymbol{O}_t^2, ..., \boldsymbol{O}_t^h\right)
\end{aligned} \tag{3,4}$$

In formula (3), the superscript $i$ denotes the head index in MSR, and $X \in \mathbb{R}^{B \times N \times W \times d_{in}}$ is the input to the $i$-th head of the MSR during parallel training. $W_Q^i \in \mathbb{R}^{d_{in} \times d}$, $W_K^i \in \mathbb{R}^{d_{in} \times d}$, and $W_V^i \in \mathbb{R}^{d_{in} \times d_{h\_out}}$ are learnable parameters used to obtain the projection matrices: $Q^i \in \mathbb{R}^{B \times N \times W \times d}$, $K^i \in \mathbb{R}^{B \times N \times W \times d}$, and $V^i \in \mathbb{R}^{B \times N \times W \times d_{h\_out}}$. Here, $d_{in}$ represents the feature dimension at each time point, $d$ is the dimension after the query and key projections, and $d_{h\_out}$ is the dimension after the value projection, which is also the output feature dimension at each time point. $\Theta$ and $\bar{\Theta}$ are conjugate matrices used for positional encoding. $D \in \mathbb{R}^{W \times W}$ is a lower triangular matrix used to mask the results of $Q^i K^{iT}$, indicating causal relationships. The calculation of $\gamma$ in $D$ follows the method described in reference (Sun, Dong, Huang, Ma, Xia, Xue, Wang & Wei, 2023), specifically $\gamma^i = 1 - 2^{-5-\text{arange}(0,h)} \in \mathbb{R}^h$, which remains consistent across different RetNet layers. After the outputs of the $h$ heads are concatenated, $GN_h$ is applied to normalize each head's result, yielding the final output $O^i \in \mathbb{R}^{B \times N \times W \times d_{out}}$, where $d_{out} = d_{h_{out}} \times h$.

In formula (4), the subscript $t$ denotes the time step, and $X_t \in \mathbb{R}^{B \times N \times 1 \times d_{in}}$ represents the input to each head at time $t$ during serialized inference. $W_Q^i$, $W_K^i$, and $W_V^i$ have the same meanings as in formula (3) and share the weights trained during the training phase. $S_t^i \in \mathbb{R}^{d \times d_{out}}$ represents the state retention matrix at time $t$, which is derived from the state retention matrix $S_{t-1}^i$ at the previous time step and the state $K_t^{iT} V_t^i$ at time $t$. After the results from $h$ heads are concatenated, the output at time $t$ is obtained as $O_t^i \in \mathbb{R}^{B \times N \times 1 \times d_{out}}$.

*4.3. Intermodal correlation extraction module*

To extract latent intermodal correlation information from the data, this paper introduces CR block in the original RetNet architecture. In the redesigned RetNet structure, the CR block is positioned after the MSR block, as illustrated in Fig. 1. The CR block integrates intermodal features at the modal scale, with the input tensor dimensions being batch, node, modality, and sequence. Similar to the MSR block, the CR block has different forward computation processes during the training and inference phases.

During parallel training, the CR block splits the query, key, and value matrices according to modality. It then computes the interaction between the query vector $Q^i$ of modality $i$ and the key vectors $K^j$ of all other modalities where $j \neq i$. This cross-modality computation establishes correlations between different modalities, allowing the CR block to capture the intermodal relationships in multimodal data. The forward computation process of the CR block during parallel training can be described by formula (5).

$$\begin{aligned}
&\boldsymbol{Q}^i = \boldsymbol{X}^i \boldsymbol{W}_Q^i \odot \boldsymbol{\Theta}, \boldsymbol{K}^i = \boldsymbol{X}^i \boldsymbol{W}_K^i \odot \bar{\boldsymbol{\Theta}}, \boldsymbol{V}^i = \boldsymbol{X}^i \boldsymbol{W}_V^i \\
&\boldsymbol{\Theta} = e^{it\theta}, \boldsymbol{D}_{t_1 t_2} = \begin{cases} \gamma^{t_1 - t_2}, & t_1 \geq t_2 \\ 0, & t_1 < t_2 \end{cases} \\
&\boldsymbol{O}^i = \left(\left(\sum_{j \neq i} \boldsymbol{Q}^i \boldsymbol{K}^{jT}\right) \odot \boldsymbol{D}\right) \boldsymbol{V}^i \\
&\boldsymbol{O} = GN_M\left(\text{concat}\left(\boldsymbol{O}^1, \boldsymbol{O}^2, ..., \boldsymbol{O}^M\right)\right)
\end{aligned} \tag{5}$$

In formula (5), the superscripts $i, j \in [1, M], i \neq j$ denote modality indices. $X^i \in \mathbb{R}^{B \times N \times W \times d_i}$ represents the data corresponding to modality $i$ in the input tensor $X \in \mathbb{R}^{B \times N \times W \times d_{in}}$, where $d_{in} = \sum_{i=1}^{M} d_i$ and where $d_i$ is the feature dimension of modality $i$. $W_Q^i, W_K^i \in \mathbb{R}^{d_i \times d}$ are learnable parameters, with $d$ being the dimension of the mapped features. After mapping and applying positional encoding, the query and key vectors for each modality are obtained: $Q^i, K^i \in \mathbb{R}^{B \times N \times W \times d}$. $W_V^i \in \mathbb{R}^{d_i \times d_v}$, where $d_v$ is the feature dimension of the value vectors for each modality. The output tensor of the CR block is $O \in \mathbb{R}^{B \times N \times W \times d_{out}}$, where $d_{out} = M d_v$. $GN_M$ represents normalization applied to the outputs of each modality separately. During the forward computation of the CR module, $\gamma$ reflects the relationship between time points $t_1$ and $t_2$, with $\gamma = 1 - 2^{-5}$, which is consistent with the value in reference (Sun, Dong, Huang, Ma, Xia, Xue, Wang & Wei, 2023) and remains the same across different RetNet layers.

In the serialization inference phase, the CR first obtains the query, key, and value matrices for different modalities at the current time $t$. To compute the current state, the value vector $V^i$ of modality $i$ interacts with the key vectors $K^j$ of all other modalities $j \neq i$ to capture the interactions between different modalities. The formula for the serialization inference phase of the CR can be expressed as follows:

$$Q_t^i = X_t W_t^i \odot \Theta, K_t^i = X_t W_t^i \odot \overline{\Theta}, V_t^i = X_t W_t^i$$

$$S_t^i = \gamma S_{t-1}^i + \left(\sum_{j \neq i} K_t^j\right)^T V_t^i, \quad S_0^i = 0$$

$$O_t^i = Q_t^i S_t^i, \quad i, j \in [1, M], t \in [1, W]$$

$$O_t = GN_M\left(\text{concat}\left(O_t^1, O_t^2, ..., O_t^m\right)\right) \tag{6}$$

The learnable parameters $W_Q^i$, $W_K^i$, and $W_V^i$ in formula (6) are the same as those used in the parallel training phase. $Q_t^i$, $K_t^i$, and $V_t^i$ represent the query, key, and value vectors of modality $i$ at time $t$, respectively. The current state is obtained by applying the value vector $V_t^i$ of the current modality to the key vectors $K_t^j$ of other modalities at the same time. This result is then added to the previous state retention matrix $S_{t-1}^i$ to produce the current state retention matrix $S_t^i$. The initial state retention matrix $S_0^i$ for each modality is set to a zero matrix. By concatenating the output vectors $O_t^i$ of all the modalities at time $t$ and applying GroupNorm for normalization, the CR block produces the output $O_t$ corresponding to $X_t$.

For the task of detecting anomalous nodes in WSNs, the input tensor $X$ in the CR block has identical dimensions across all modalities. This allows parallel computation of each modality to increase training efficiency. Taking the training phase as an example, the process of fusing other modal features can be represented in the following parallelized form:

$$W_Q = concat\left(\left\{W_Q^i \mid i \in [1, M]\right\}\right)$$

$$W_K = concat\left(\left\{W_K^i \mid i \in [1, M]\right\}\right)$$

$$W_V = concat\left(\left\{W_V^i \mid i \in [1, M]\right\}\right)$$

$$Q = XW_Q \odot \Theta, K = XW_K \odot \overline{\Theta}, V = XW_V$$

$$K_{cross} = sum(K, \dim = 2) - K$$

$$O = \left(QK_{cross}^T \odot D\right)V$$

$$O_{out} = GN_M\left(reshape(O)\right) \tag{7}$$

In formula (7), the input $X^i$ for each modality is merged into $X \in \mathbb{R}^{B \times N \times M \times W \times d_i}$. For example, considering the query mapping computation, $W_Q \in \mathbb{R}^{M \times d_i \times d}$ for $i \in [1, M]$ is combined into $W_Q \in \mathbb{R}^{M \times d_i \times d}$. The tensors $X$ and $W_Q$ are used in a broadcasting mechanism to obtain the mapping matrix $Q \in \mathbb{R}^{B \times N \times M \times W \times d}$. Similarly, $K \in \mathbb{R}^{B \times N \times M \times W \times d}$ and $V \in \mathbb{R}^{B \times N \times M \times W \times d_v}$ are computed. The term $K_{cross}$ denotes the sum of key mappings from other modalities, which is equivalent to $\sum_{j \neq i} K_j$ in formula (5). This method yields $O \in \mathbb{R}^{B \times N \times M \times W \times d_v}$, which is then reshaped and normalized to produce the final output $O \in \mathbb{R}^{B \times N \times W \times d_{out}}$.

*4.4. Multigranularity feature fusion module*

To obtain node feature representations, this paper employs a multigranularity feature fusion module to integrate various granularity and modality features within each node, as illustrated in Fig. 1. The feature pyramid network (FPN) structure is positioned after the improved RetNet. This module is designed with a multigranularity feature fusion module consisting of two linear layers. The linear layers can be represented as follows:

$$O = \text{ReLU}(XW + b) \tag{8}$$

The first linear layer has $W \in \mathbb{R}^{dL \times d}, b \in \mathbb{R}^d$, with the fused result represented as $O \in \mathbb{R}^{B \times N \times W \times d}$. This layer is used to fuse the different fine-grained features obtained from each RetNet layer, which helps to enrich feature representations and enhance the model's expressive capability.

During the training phase, the outputs of each RetNet layer are stacked and used as input to the linear layer, denoted as $X \in \mathbb{R}^{B \times N \times W \times dL}$, where L is the number of layers in the improved RetNet structure. In the inference process, the multigranularity feature fusion module combines the inference results from the $t$-th moment and the previous $W - 1$ moments of the L layers of RetNet as inputs to the linear layer, represented as $X \in \mathbb{R}^{B \times N \times W \times dL}$. The subsequent computation process is the same as that in the training phase. Therefore, for the module following this linear layer, the forward computation process during training and inference is identical.

The second linear layer of the multigranularity feature fusion module is used to merge features of different modalities within the node, effectively mapping $X \in \mathbb{R}^{B \times N \times d \times M} \to X \in \mathbb{R}^{B \times N \times d}$.

### 4.5. Spatial correlation feature extraction module

In the MTAD-RD model framework shown in Fig. 1, this paper utilizes the GAT after the FPN structure to extract correlation information between nodes in WSNs. The GAT integrates node features at the node scale, where the dimensions of the input tensor are as follows: batch, nodes, and node attributes. For a graph in a batch, let the attribute matrix be $X \in \mathbb{R}^{N \times d_{in}}$ and the adjacency matrix be $A \in \mathbb{R}^{N \times N}$. The output after applying the GAT is $X' \in \mathbb{R}^{N \times d_{out}}$, where $d_{in}$ and $d_{out}$ represent the dimensions of the input and output node feature vectors in the GAT module, respectively.

$$e_{i,j} = LeakyReLU\left(a^T \left[Wh_i \| Wh_j\right]\right)$$

$$\alpha_{i,j} = \frac{\exp(e_{i,j})}{\sum_{k \in \mathcal{N}(i)} \exp(e_{i,j})}$$

$$h_i' = \sigma\left(\sum_{j \in \mathcal{N}(i)} \alpha_{i,j} W h_j\right) \quad (9)$$

In formula (9), $a \in \mathbb{R}^{2d_{out}}$, $W \in \mathbb{R}^{d_{in} \times d_{out}}$ are learnable parameters used to map the input features. The parameter $a$ is used to compute attention scores, denoted as $e_{i,j}$. After normalizing the attention scores via $softmax$, we obtain the attention coefficients. The attention coefficient between node $i$ and node $j$ is $\alpha_{i,j}$. Consequently, the feature representation of node $i$, after integrating information from neighboring nodes, is $h_i' \in \mathbb{R}^{d_{out}}$.

### 4.6. Unsupervised pretraining of the backbone network

Considering the lack of labeled data in practical application scenarios, this paper employs a contrastive learning approach to design a pretraining method for the backbone network to extract transferable features from unlabeled data in the WSN dataset.

This paper introduces a proxy task of contrasting nodes with subgraphs. For an attribute graph $G = (X, A)$, where $X \in \mathbb{R}^{N \times d}$ and $A \in \mathbb{R}^{N \times N}$, we first randomly select a node $N_a$ from the node set $N$ and use its features as an anchor sample, denoted as $n_a$. We then sample subgraphs from other nodes in $G$. The sampling method for the subgraphs corresponding to node $N_i$ is as follows:

$$S_i = \text{RandomSubgraph}(N_i, k) + \text{Neighbor}(N_i) \quad (10)$$

In formula (10), $S_i$ represents the subgraph corresponding to node $N_i$, which consists of two parts: $RandomSubgraph(N_i, k)$, which denotes a random walk-based sampling of $k$ nodes starting from nodes $N_i$, and $Neighbor(N_i)$, which represents the neighbors of node $N_i$. Including $Neighbor(N_i)$ in $S_i$ helps prevent the loss of potentially anomalous correlation information between node $N_i$ and its neighbors. The subgraph $S_i$ is constructed by combining these two sets of nodes, and the subgraph features $s_i$ are obtained by applying $AveragePooling$ to the features of nodes in $S_i$. We denote the set of $s_i$ as $S$.

In this work, the Pearson correlation coefficient is used to calculate the correlation between anchor samples and each subgraph. This correlation is then used to construct positive and negative sample pairs. The calculation process is as follows:

$$\rho(n_a, s_i) = \frac{cov(n_a, s_i)}{\sigma(n_a)\sigma(s_i)} \quad (11)$$

In formula (11), $cov(n_a, s_i)$ denotes the covariance between $n_a$ and $s_i$, whereas $\sigma(n_a)$ and $\sigma(s_i)$ represent the standard deviations of $n_a$ and $s_i$, respectively. To calculate the Pearson coefficient between each sample and the anchor sample, the sample with the highest correlation is selected as the positive sample for $n_a$, denoted as $s_p$. The $k$ samples with the lowest correlations are chosen as negative samples, and the set of negative samples is denoted as $S_n$.

The InfoNCE loss $\mathcal{L}_{NCE}$ is then computed via the anchor sample $n_a$, the positive sample $s_p$, and the negative sample set $S_n$. The optimization objective is to minimize $\mathcal{L}_{NCE}$. The calculation process for $\mathcal{L}_{NCE}$, with $\tau$ representing the temperature parameter, is designed as follows:

$$\mathcal{L}_{NCE} = -\log \frac{exp(n_a \cdot s_p / \tau)}{\sum_{i \in \mathcal{N}(n)} exp(n_a \cdot s_i / \tau)} \quad (12)$$

### 4.7. Anomaly discriminant network

On the basis of the dual-graph neural network architecture (Yang, Li, Zhang, Zhou, Zhou & Liu, 2020; Wang, Wang, Zhao, Liu, Liu & Shen, 2023), this paper designs the MTAD-RD model's anomaly node classification network. Unlike previous methods reported in the literature, this paper integrates contrastive learning with few-shot learning to jointly train the anomaly node classification network in a weakly supervised manner. During the graph embedding propagation process, nodes continuously aggregate information from their neighbors. Leveraging this characteristic, the anomaly node classification network combines the limited label information from the dataset with the WSN node features extracted by the backbone network as input, while unlabeled data are assigned a label of 0. Data in the same batch are organized into instance graphs and distribution graphs in the form of a complete graph. During the alternating update process of these graphs, nodes with real label information propagate their labels to other nodes. After L rounds of alternating updates, the model can predict its own class on the basis of the labels of neighboring nodes and the weights of the edges connecting them. The training process for the designed classification network involves three steps: sampling and initialization, feature embedding propagation, and joint optimization. The specific design of these steps is as follows:

**Sampling and initialization:** In typical WSN datasets, there is often an imbalance between the number of normal samples and anomalous samples, with anomalous samples being less frequent. To address this, the paper first sets up a buffer to store anomalous samples from each batch via a first-in-first-out (FIFO) method. Next, it performs balanced sampling on the feature set $X_{raw} \in \mathbb{R}^{B \times N \times d}$ of the current training batch, obtaining K normal nodes $X_{normal} \in \mathbb{R}^{K \times d}$ and K anomalous nodes $X_{abnormal} \in \mathbb{R}^{K \times d}$. If there are not enough anomalous nodes in the current batch, it retrieves the most recent anomalous samples from the buffer. The unsampled labeled nodes in this batch are denoted as $X_{label}$, and all unlabeled nodes are denoted as $X_{unlabel}$. From the perspective of few-shot learning, $X_{normal}$ and $X_{abnormal}$ together form the support set, whereas $X_{label}$ and $X_{unlabel}$ collectively constitute the query set. From the perspective of contrastive learning, one node from the normal nodes is selected as the anchor sample, denoted as $n_a$. The remaining normal samples form the positive sample set, denoted as $X_{pos}$, and all anomalous samples form the negative sample set, denoted as $X_{neg}$. The anomalous node detection network uses WSN node features and their one-hot labels as node features for the instance graph, with the samples in the query set having zero vector labels.

The initialization process for the instance graph and distribution graph in the anomalous node detection network is as follows:

$$\begin{aligned}
v_{0,i}^{ins} &= \text{concat}(x_i, y_i) \\
e_{0,ij}^{ins} &= \text{MLP}_0^{ins}\left(\left(v_{0,i}^{ins} - v_{0,j}^{ins}\right)^2\right) \\
v_{0,i}^{dis} &= \begin{cases} \|\sigma(y_i, y_j)\| & v_i \in \text{SupportSet} \\ \dfrac{1}{NK}[1,1,...,1] & v_i \in \text{QuerySet} \end{cases} \\
e_{0,ij}^{ins} &= \text{MLP}_0^{dis}\left(\left(v_{0,i}^{ins} - v_{0,j}^{ins}\right)^2\right)
\end{aligned} \quad (13)$$

In formula (13), $v_{0,i}^{ins} \in \mathbb{R}^{d+d_l}$ represents the initial state of node $i$ in the instance graph at layer 0. Here, $x_i$ denotes the feature of the WSN node obtained from the backbone network, with the feature vector length being $d$, whereas $y_i$ is the one-hot encoded label corresponding to $x_i$, with the label vector length being $d_l$. For unlabeled data, $y_i$ is a zero vector. $v_{0,i}^{dis} \in \mathbb{R}^{NK}$ represents the initial state in the distribution graph for node $i$. The initial state of $v_{0,i}^{dis}$ is determined by $x_i$ and the labels of its neighbors. If $x_i$ belongs to the support set, $v_{0,i}^{dis}$ consists of binary values of 1 and 0. If the labels $y_i$ and $y_j$ corresponding to nodes $x_i$ and $x_j$ in the instance graph are the same, then $\sigma(y_i, y_j) = 1$ and $v_{0,i}^{dis}[j] = 1$; otherwise, it is zero. If $x_i$ belongs to the query set, each value in the feature vector of the node is $1/NK$, where $N$ is the number of categories sampled in the support set and where $K$ is the number of samples per category. $e_{0,ij}^{ins} \in \mathbb{R}$ represents the weight of the edge between $v_{0,i}^{ins}$ and $v_{0,j}^{ins}$, with both node feature vectors processed through an MLP network to obtain $e_{0,ij}^{dis} \in \mathbb{R}$, which has a similar meaning.

**Feature Embedding Propagation:** In the anomaly detection network, the instance graph and distribution graph are updated alternately. Specifically, the edges of the instance graph, the nodes of the distribution graph, the edges of the distribution graph, and the nodes of the instance graph are updated sequentially. Each update process is handled by different MLP networks, as represented by the following formulas:

$$\begin{aligned}
e_{l,ij}^{ins} &= \text{MLP}_l^{ins}\left(\left(v_{l-1,i}^{ins} - v_{l-1,j}^{ins}\right)^2\right) \cdot e_{l-1,ij}^{ins} \\
v_{l,i}^{ins} &= \text{MLP}_l^{d2i}\left(v_{l-1,i}^{ins}, \sum_{j \neq i}\left(e_{l,ij}^{dis} \cdot v_{l-1,i}^{ins}\right)\right) \\
v_{l,i}^{dis} &= \text{MLP}_l^{i2d}\left(v_{l-1,i}^{dis}, \|e_{l,ij}^{ins}\|\right) \\
e_{l,ij}^{dis} &= \text{MLP}_l^{dis}\left(\left(v_{l-1,i}^{dis} - v_{l-1,j}^{dis}\right)^2\right) \cdot e_{l-1,ij}^{dis}
\end{aligned} \quad (14)$$

In formula (14), $l \in [1, L]$ indicates that different MLP networks are used in different update rounds. The functions of $MLP_l^{ins}$ and $MLP_l^{dis}$ are similar to those of $MLP_0^{ins}$ and $MLP_0^{dis}$ in formula (13). The difference lies in the fact that the calculation of $e_{l,ij}^{ins}$ and $e_{l,ij}^{dis}$ involves the edges from the previous layer. The MLP network $MLP_l^{d2i}$ is used to update the instance graph nodes, taking data from two sources: one part from the instance graph nodes $v_{l-1,i}^{ins}$ from the previous layer and the other part from the aggregated information of other instance graph nodes $v_{l-1,i}^{ins}$ connected by the edges $e_{l,ij}^{dis}$ in the same layer.

**Joint Optimization:** The anomaly detection network computes the probability of each labeled node belonging to various classes through the instance graph and the distribution graph. The calculation method is designed as follows:

$$\begin{aligned}
\hat{y}_{l,i}^{ins} &= \text{softmax}\left(\lambda_1 \sum_{j=1}^{NK} e_{l,ij}^{ins} \cdot y_j + (1-\lambda_1)\hat{y}_i\right) \\
\hat{y}_{l,i}^{dis} &= \text{softmax}\left(\lambda_2 \sum_{j=1}^{NK} e_{l,ij}^{dis} \cdot y_j + (1-\lambda_2)\sum_{j=1}^{NK} v_{l,i}^{dis}[j] \cdot y_j\right)
\end{aligned} \quad (15)$$

In this context, $\lambda_1$ and $\lambda_2$ are coefficients, $y_i$ represents the one-hot label corresponding to sample $x_i$ in $X_{label}$, and $\hat{y}_1$ is the result after mapping the instance graph for $y_i$, i.e., $x_i y_i \to \hat{x}_i \hat{y}_i$. $y_j$ is the one-hot label corresponding to sample $x_j$ in the support set, and $e_{l,ij}^{ins}$ denotes the weight between the node corresponding to $x_i$ and the node corresponding to $x_j$ in the $l$-th layer of the instance graph. The prediction of the class information for $x_i$ in the $l$-th layer instance graph consists of two parts: one part aggregates the label information of neighbors through the instance graph edge $e_{l,ij}^{ins}$, and the other part is the information derived from the instance graph node $\hat{y}_l$. These two prediction components are used to constrain the update processes of the instance graph edges and nodes during optimization. $v_{l,i}^{dis}$ represents the feature representation of the $x_i$ node in the $l$-th layer distribution graph, reflecting the relationship between $x_i$ and $x_j$, whereas $e_{l,ij}^{dis}$ is the weight between the node corresponding to $x_i$ and the node corresponding to $x_j$ in the $l$-th layer distribution graph, reflecting the relationship between $x_i$ and $x_j$. The prediction of the class information for $x_i$ in the $l$-th layer instance graph is obtained by calculating $e_{l,ij}^{dis}$ and $v_{l,i}^{dis}$ with $y_i$. These prediction components are used to constrain the update processes of the distribution graph edges and nodes during optimization. The loss functions for the distribution graph and instance graph in the proposed optimization process are calculated as follows:

$$\mathcal{L}_l^{ins} = \sum_{i \in X_{\text{label}}} \text{CE}\left(\hat{\boldsymbol{y}}_{l,i}^{ins}, \boldsymbol{y}_i\right)$$
$$\mathcal{L}_l^{dis} = \sum_{i \in X_{\text{label}}} \text{CE}\left(\hat{\boldsymbol{y}}_{l,i}^{dis}, \boldsymbol{y}_i\right) \quad (16)$$

During the training of the anomaly node detection network, contrastive learning methods are also employed. The contrastive loss function is defined as follows:

$$\mathcal{L}_{cont} = -\log \frac{\sum_{v_i \in X_{\text{normal}}} \boldsymbol{n}_a \cdot \boldsymbol{v}_i / \tau}{\sum_{v_j \in X_{\text{abnormal}}} \boldsymbol{n}_a \cdot \boldsymbol{v}_j / \tau} \quad (17)$$

$$\mathcal{L} = \omega \mathcal{L}_{cont} + (1-\omega) \sum_{l=1}^{L} \frac{1}{2^{L-l}} \left(\mathcal{L}_l^{dis} + \mathcal{L}_l^{dis}\right) \quad (18)$$

In formula (17), $\tau$ is the temperature coefficient, $n_a$ represents anchor samples, $v_i$ denotes positive samples, and $v_j$ denotes negative samples. The joint loss function for optimizing the anomaly node detection network is given in formula (18), where $L$ represents the number of layers in the instance and distribution graphs, and $\omega$ is a hyperparameter used to adjust the proportion of contrast loss in joint loss. This paper discusses the influence of the proportion of $\mathcal{L}_{cont}$ in joint loss on the training effect of the model in the subsequent comparison experiment.

## 5. Experiments and Analysis

### 5.1. The dataset and experimental setup

The dataset used in this paper was collected from a real-world deployment at the Intel Berkeley Research Laboratory (IBRL). This dataset contains nearly two months of data, including humidity, temperature, light, and voltage readings from 54 sensors. The actual locations of the sensors are shown in Fig. 2.

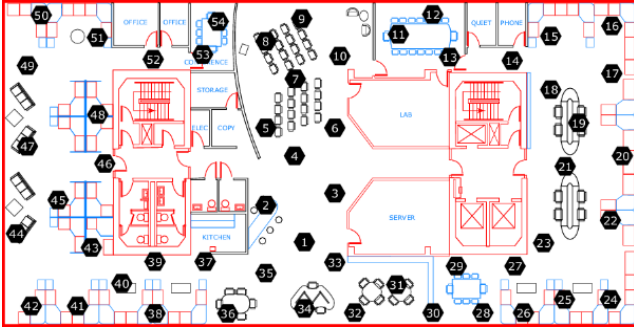

Fig. 2. Spatial distribution of sensor nodes in the IBRL dataset.

The original IBRL dataset has a significant amount of missing data. Upon analysis, we found that nodes 5, 15, and 18 had substantial missing values, particularly for the light modality across most sensors. Therefore, in our study, we excluded these three nodes. Ultimately, we used humidity, temperature, and voltage data from 51 sensor nodes for our experiments. Owing to network transmission issues, the data reporting times were not synchronized. To address this, we aligned the timestamps by segmenting the entire period on the basis of a fixed sampling interval. For each interval, the average of multiple data points was used as the value at the start of that interval, with the value closest to the sampling time being selected as the final sampling value. If no data points were available within a given interval, we performed weighted interpolation using the two closest data points, with the weights determined by their distance to the start of the interval. The processed dataset was then split into training set, verification set and test set at a 7:2:1 ratio.

In the selected period, there were no significant anomalies in the data. Therefore, we artificially injected point anomalies, collective anomalies, contextual anomalies, and two types of correlation anomalies into the IBRL dataset. We followed the anomaly injection method proposed by Zhang et al. (Ye, Zhang, Xue, Wang, Jiang & Qiu, 2024), with an anomaly sample injection rate of 2%. After injection, the dataset contained 3% labeled samples, with the ratio of labeled normal samples to labeled anomaly samples being 2:1.

The hardware platform for our experiments consisted of an Intel Xeon Silver 4210 CPU @ 2.20 GHz and an NVIDIA GeForce RTX 2080Ti GPU. The software environments used were Ubuntu 20.04, Python 3.8, and PyTorch 1.13.1+cu117. In subsequent experiments, we employed a three-layer RetNet architecture for the MTAD-RD, alternately updating the instance and distribution graphs in the discriminator five times. The batch size was set to 16, the learning rate was 0.0001, and the model was optimized via the Adam optimizer.

### 5.2. Evaluation metrics

In this study, the precision, recall, and F1 score are used as evaluation metrics for model performance. Precision measures the accuracy of the model when it is used to predict positive (Positive) instances. It is defined as the ratio of correctly predicted positive instances to the total number of instances predicted as positive. Precision reflects the proportion of true positives among all instances predicted as positive. High precision indicates that the model makes few false positive predictions when identifying positive instances. The formula for precision is as follows:

$$\text{Precision} = \frac{\text{TP}}{\text{TP+FP}} \quad (19)$$

Recall, also known as sensitivity, that the model's ability to correctly identify all actual positive instances is evaluated. A high recall means that the model captures more positive instances, although it may lead to more false positives. The formula for recall is as follows:

$$\text{Recall} = \frac{\text{TP}}{\text{TP+FN}} \quad (20)$$

In the above formulas, TP represents true positives (positive samples correctly predicted by the model), TN denotes true negatives, FP stands for false positives (negative samples incorrectly classified as positive), and FN refers to false negatives.

The F1 score is the harmonic mean of precision and recall, accounting for both the model's accuracy and its ability to recall positive instances. It strikes a balance between precision and recall, making it particularly useful when the dataset has imbalanced class distributions. The F1 score is calculated as follows:

$$F1=\frac{2\times \text{Precision}\times \text{Recall}}{\text{Precision}+\text{Recall}} \qquad (21)$$

*5.3. Ablation study*

To investigate the impact of various modules in the backbone network on anomalous node detection and the influence of unsupervised contrastive learning-based pretraining on the overall performance of the model, we designed a set of ablation experiments for the proposed model. By removing or modifying different modules in the model, we can assess the contribution of each module to the overall performance. The following ablation configurations were designed:

**Experiment 1:** Use the original RetNet without separately extracting cross-modal correlation information.

**Experiment 2:** Remove the GAT module from the backbone network, avoiding the separate extraction of node correlation information.

**Experiment 3:** Remove both the CR and GAT modules, thus not extracting any correlation information.

**Experiment 4:** Exclude the FPN structure for multiscale feature fusion and instead directly feed the output from the final layer of RetNet into the anomalous node detection network.

**Experiment 5:** Use only one layer of the modified RetNet in the backbone network.

**Experiment 6:** Cancel unsupervised contrastive learning pretraining for the feature extraction network and optimize both the backbone and detection networks simultaneously in a weakly supervised manner using only limited labeled data while keeping the number of training epochs unchanged.

**Experiment 7:** Based on Configuration 6, increase the training epochs to 300, also, make sure that the weights of the backbone network are frozen during the final 100 epochs.

**Experiment 8:** Train the complete model in two stages. The backbone network consists of three layers of the modified RetNet, with the backbone pretrained for 200 epochs via unsupervised contrastive learning, followed by 100 epochs of joint training of the detection network via contrastive learning and few-shot learning methods, which freeze the backbone weights for the final 70 epochs.

Table 2

Ablation Study Designs and Results

| No. | CR | FPN | GAT | Pretraining | Precision | Recall | F1 | Description |
|---|---|---|---|---|---|---|---|---|
| 1 | × | √ | √ | √ | 92.53% | 72.55% | 81.33% | Not using CR to extract intermodal correlations individually. |
| 2 | √ | √ | × | √ | 92.27% | 71.17% | 80.36% | Not using GAT to extract internode correlations individually. |
| 3 | × | √ | × | √ | 72.73% | 53.10% | 61.38% | Not handling correlation information separately. |
| 4 | √ | × | √ | √ | 82.23% | 85.33% | 83.75% | Not using the Feature Extraction Module. |
| 5 | √ | × | √ | √ | 67.20% | 75.05% | 70.91% | Using a single-layer improved RetNet in the backbone network. |
| 6 | √ | √ | √ | × | 62.40% | 68.33% | 65.23% | Eliminating unsupervised pretraining of the backbone network. |
| 7 | √ | √ | √ | × | 86.33% | 88.67% | 87.48% | Increase the number of training epochs based on Scheme 6. |
| 8 | √ | √ | √ | √ | 89.87% | **92.10%** | **90.97%** | The model designed in this paper. |

The results of the ablation experiments are presented in Table 2. In Configuration 1, the CR module in the backbone network is disabled, leading to a 19.55% decrease in recall and a 9.64% decrease in the F1 score compared with those of the baseline. This is primarily due to a significant number of missed detections of cross-modal correlations. In Configuration 2, the GAT module is disabled, preventing WSN nodes from integrating information from neighboring nodes. As a result, the recall decreases by 20.93, and the F1 score decreases by 10.61% compared with the baseline. This is caused by the model's degraded performance in detecting internode correlations. Interestingly, both Configuration 1 and Configuration 2 achieve higher precision than the baseline does, despite missing correlation anomalies, indicating that the current dataset poses a greater challenge for the baseline in detecting correlation anomalies. Additionally, the model performs better in detecting point, collective, and contextual anomalies than in detecting correlation anomalies. In Configuration 3, where both the CR and GAT modules are disabled, the recall decreases by 39%, and the precision decreases by 17.17%. This suggests that relying solely on the MSR module to extract temporal features is insufficient to fully utilize the latent information in the WSN dataset.

To gain further insight into the specific effects of different loss weights in the joint loss function on the training of the model, a series of experimental schemes were designed, which are presented in Table 3. In the five schemes (A to E), the value of the hyperparameter in formula (18) was increased from 0.2 to 0.8, whereas in scheme F, the contrastive loss was removed from the joint loss function.

The results of the experiments conducted with different joint loss function design schemes are presented in Table 3. As the ω values of schemes A to F increased from 0 to 0.8, the recall rate and F1 score of the model fluctuated. The model attained its optimal performance when ω was set to 0.4. Compared with those in Scenario A, in which only instance graph loss and distribution graph loss were employed, the recall increased by 4.37% and the F1 score increased by 5.14% in Scenario C following the incorporation of contrastive loss. The efficacy of the discriminant network in detecting anomalies may be increased by constraining the distribution of nodes in the feature space through contrastive loss. As the value of ω increased from 0.4 to 0.8, the recall and F1 score of the model decreased. This may be attributable to reductions in the percentages of instance graph loss and distribution graph loss, which would in turn affect the transfer of labeled information. An increase in the proportion of contrastive loss within the joint loss function is beneficial for constraining the distribution of features among nodes within an instance graph in a feature space. However, an excessively high proportion of contrastive loss results in a weakening of the constraining effect of the joint loss function on edges within both the instance and distribution graphs during the updating process. This may cause the nodes in the instance and distribution graphs to struggle with efficiently aggregating labeling information from their neighboring nodes via the connecting edges.

Table 3

Joint loss function design schemes and results

| No. | $\omega$ | Precision | Recall | F1 Score |
|---|---|---|---|---|
| A | 0 | 85.33% | 86.33% | 85.83% |
| B | 0.2 | 88.33% | 91.33% | 89.80% |
| C | 0.4 | **89.87%** | **92.10%** | **90.97%** |
| D | 0.5 | 87.48% | 91.73% | 89.55% |
| E | 0.6 | 82.33% | 72.73% | 77.23% |
| F | 0.8 | 72.73% | 68.33% | 70.46% |

*5.4. Comparison experiment*

This section compares the proposed MTAD-RD model with CNN-LSTM, MTAD-GAT, GAT-GRU, and GLSL. Experiments were conducted on the IBRL dataset to evaluate the performance of these models in terms of recall, precision, and F1 score. The experimental results are shown in Table 4, where the MTAD-RD represents the model proposed in this paper.

**CNN-LSTM** (Zeng, Chen, Qian, Wang, Zhou & Tang, 2023): This method uses a deep learning network constructed with convolutional modules, LSTM modules, and fully connected layers. The convolutional module employs a 2D convolution block with a kernel size of 3. The output of the convolutional module is fed into the LSTM, and the final classification output is obtained through fully connected layers.

**MTAD-GAT** (Zhao, Wang, Duan, Huang, Cao, Tong, Xu, Bai, Tong & Zhang, 2020): This anomaly detection method combines the strengths of both predictive and reconstructive models. The predictive component models the temporal dependencies of the data, capturing subtle anomalous changes that assist in the reconstruction process. The MTAD-GAT network consists of two GATs: the first treats each time series as a node, capturing inter-series correlations, while the second considers the data at different time steps as nodes, capturing temporal dependencies. The features learned from these two GATs are combined with the original data and fed into a GRU to capture long-term dependencies. The model predicts and reconstructs the features of each time series, and by combining prediction and reconstruction errors, it can more accurately identify anomalous data points, increasing sensitivity and detection accuracy.

**GAT-GRU** (Zhang, Ye & Deng, 2022): This is a network structure capable of handling multi-node, multimodal time series data. The model first creates feature extraction branches for each node in the WSNs, using GAT to extract correlation information between different modalities within each node. It then applies the GAT to aggregate features from neighboring nodes, capturing internode correlation information. Finally, the GRU is used to process the features of each node, extracting temporal information. The authors perform anomaly detection via a reconstruction-based method, where data are mapped to a lower-dimensional space via GAT-GRU and then restored to its original high-dimensional space via a similar structure. The reconstruction loss is used as the anomaly detection score.

**GLSL** (Ye, Zhang, Xue, Wang, Jiang & Qiu, 2024): This method is an improvement over GAT-GRU. Like GAT-GRU, GLSL uses two distinct GAT modules to extract both node and intermodal correlations from WSN data. The key difference is that GLSL creates feature extraction branches for different modalities, and altering the computation order of the two GAT modules prevents the number of branches from increasing as the WSN scale increases.

Precision measures the proportion of true positives among all samples predicted as positive, which is directly correlated with the design and training of the discriminative network. As shown in Table 4, the precision of the proposed method is 3.43% and 4.63% lower than those of the GAT-GRU and GLSL methods, respectively, both of which are based on reconstruction models. This gap arises mainly because the discriminative network in our method is trained in a weakly supervised manner, whereas GAT-GRU and GLSL leverage labeled normal data during training. Nevertheless, our method outperforms the supervised CNN-LSTM and MTAD-GAT methods in terms of anomaly detection precision.

Table 4

Results of Comparative Experiments

| Method | Precision | Recall | F1 Score | MFLOPs |
|---|---|---|---|---|
| CNN-LSTM | 79.5% | 70.0% | 74.5% | 124.830 |
| MTAD-GAT | 77.5% | 86.0% | 80.5% | 21953.475 |
| GAT-GRU | 93.3% | 87.5% | 90.3% | 258.264 |
| GLSL | <u>94.5%</u> | 87.0% | 90.6% | 203.829 |
| MTAD-RD | 89.87% | **92.10%** | **90.97%** | **86.723** |

Recall measures the proportion of true positives among all actual positive samples and is not only tied to the anomaly detection capability of the discriminative network but also closely related to the feature extraction ability of the backbone network. According to the results in Table 4, the CNN-LSTM model uses convolutional kernels to capture correlations across multiple time series. However, owing to the limited size of the convolutional kernels, it cannot capture relationships across different time series simultaneously, which limits its performance. As a result, its recall is 22.10% lower than that of our method. MTAD-GAT improves on this by using GAT to capture temporal correlations, leading to a significant recall improvement—16% greater than that of CNN-LSTM. However, extracting correlations across all time series directly in the MTAD-GAT can result in noisy information, where correlations between nodes and modalities interfere with each other, thus affecting the model's performance. GAT-GRU and GLSL address this by separating correlations into node-level and modality-level correlations, with distinct GAT blocks handling each. This approach more accurately captures dependencies between nodes and within modalities in WSNs. Compared with the MTAD-GAT, GAT-GRU and GLSL improve the recall by 1.5% and 1%, respectively.

Our proposed MTAD-RD embeds CR blocks in RetNet to capture modality-level correlations and uses an FPN to fuse information extracted by RetNet at multiple granularities before employing a GAT to learn node-level correlations. As a result, the recall of our method surpasses that of GAT-GRU by 5.6%.

The F1 score is the harmonic mean of precision and recall and is commonly used to provide a comprehensive evaluation of model performance. As shown in the experimental results in Table 4, owing to the high recall achieved by an effective feature extraction network, the F1 score of the proposed method is slightly higher than those of GAT-GRU and GLSL. This indicates that the model designed in this paper demonstrates superior anomaly node detection performance when trained with the combined approach of contrastive learning and few-shot learning, matching the performance of mainstream models trained with reconstruction-based methods.

MFLOPs (million floating point operations) refer to the number of million floating point operations required for a single forward pass and are typically used to measure the computational complexity of a model. The results in the last column of Table 4 show that, compared with CNN-LSTM, MTAD-GAT, GAT-GRU, and GLSL, the proposed model has the lowest computational complexity during inference, requiring only 86.723 MFLOPs per inference. Both CNN-LSTM, GAT-GRU, GLSL, and the proposed method rely on serialized inference. Compared with the feature extraction network in our method, the CNN-LSTM model has greater time complexity because of the larger number of hidden variables used by the LSTM in its forward pass. GAT-GRU and GLSL replace LSTM with GRU, reducing the complexity of temporal feature extraction. However, compared with CNN-LSTM, GAT-GRU and GLSL introduce additional processes to capture temporal correlations and implement separate branches for each node or modality, leading to greater time complexity. MTAD-GAT uses an attention mechanism to capture temporal information globally, requiring the calculation of temporal correlations at each time step, resulting in significantly higher computational complexity. When the sequence length is 200, MTAD-GAT requires 21,953.475 MFLOPs during inference. During training, MTAD-RD operates similarly to MTAD-GAT, with each time step processed in parallel. However, during inference, the MTAD-RD, like CNN-LSTM, GAT-GRU, and GLSL, uses serialized inference across time steps, which leads to a substantial reduction in computational complexity compared with parallel inference.

*5.5. Visualization Analysis*

This subsection presents a case study on the model's anomaly detection performance using a portion of the test data. Fig. 3 shows a segment of temperature data, consisting of 9,000 time points, collected by Node 1 in the IBRL dataset. The red line represents the original normal data, the green line indicates injected point anomalies, the black line corresponds to injected collective anomalies, and the blue line denotes injected contextual anomalies. In Fig. 4, the black dashed line indicates the labels for each time point in the test data. A sliding window protocol with a window size of 300 was used to construct the test samples. The results show that the detected anomalies closely align with the corresponding labels, with misclassifications occurring primarily at the edges of the sliding windows, where anomalies are just included, and in instances where the original normal data exhibited large fluctuations. Overall, the proposed anomaly detection method successfully identified the majority of the anomalies.

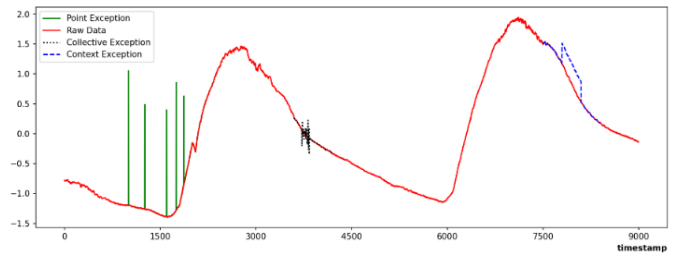

Fig. 3. Single Time Series Abnormal Test Data.

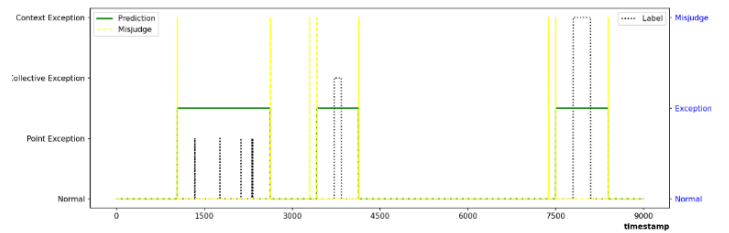

Fig. 4. Single Time Series Abnormal Test Results.

Fig. 5 shows the visualization results of testing a segment of time series data from node 27 in the IBRL dataset. Normally, the temperature time series data (red line) and humidity time series data (green line) exhibited a negative correlation. This is demonstrated in Fig. 5. However, during the time periods [1192, 1390] and [4000, 4090], the temperature and humidity time series data show an unexpected positive correlation, during the time period [2550, 3090], the correlation between the temperature and humidity time series data disappears. The figure uses black lines to denote data labels, thereby indicating the occurrence of correlation anomalies in the timing data during the three aforementioned time periods. The purple line represents the detection results from Scheme 3 in the ablation study. This scheme corresponds to a model that does not utilize CR and GAT for the fusion of correlation information. As demonstrated in Fig. 5, in the absence of CR and GAT modules, the designed model is unable to detect correlation anomalies resulting from the dissolution of inter-temporal correlations, and encounters difficulties in identifying anomalies that deviate from normal correlation patterns. Following the incorporation of the CR and GAT modules, the method outlined in this paper is capable of effectively detecting correlation anomalies, as illustrated by the blue line in Fig. 5. The orange line in the figure represents the misclassification cases of the designed model, which primarily occur at the start and end moments of anomalies caused by the loss of inter-temporal correlation. The visualisation results of the correlation anomaly detection experiment (see Fig. 5) further validate the effectiveness of the individual components of the correlation detection model designed in this paper.

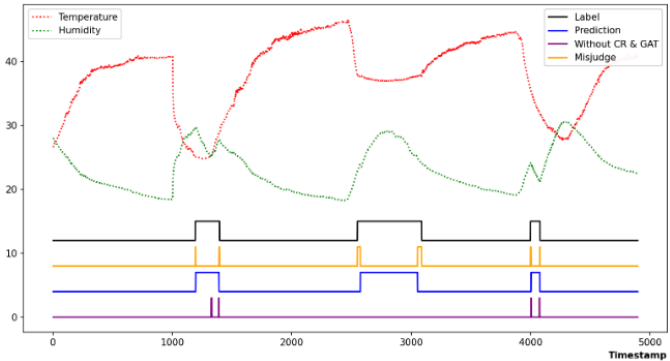

Fig. 5. Correlation Anomaly Analysis.

## 6. Summary and Prospects

In this work, we designed a multinode and multitime anomaly detection model that combines the RetNet network and a double graph discriminant network and designed a corresponding training mode for the model according to the characteristics of the WSN dataset. RetNet solves the problem of high computational complexity in the inference stage of the transformer model. We introduced RetNet into the field of WSN abnormal node detection, and through the CR module designed for it, RetNet can also extract correlation information between time series when extracting time series features. In this study, a two-stage training approach was adopted. In the first stage, unsupervised contrast learning is used to pretrain the backbone network of MTAD-GAT. By comparing nodes and subgraphs, the model can learn prior knowledge from the unlabeled data of the WSN. In the second stage, MTAD-GAT is trained with weak supervision by combining small sample learning with contrast learning. The experimental results showed that in the WSN anomaly detection task, the proposed method was obviously superior to the traditional multitime series detection model, and its anomaly detection performance was close to that of GAT-GRU and GLSL, which are based on supervised learning. In the next stage of research, we can consider how to further improve the feature extraction method, such as by extending it to the frequency domain and designing a more reasonable proxy method for comparative learning, to further improve the anomaly detection performance.